# ELKPPNet: An Edge-aware Neural Network with Large Kernel Pyramid Pooling for Learning Discriminative Features in Semantic Segmentation

Xianwei Zheng[1,*], Linxi Huan[1], Hanjiang Xiong[1], Jianya Gong[1,2]

[1]The State Key Laboratory of Information Engineering in Surveying, Mapping and Remote Sensing, Wuhan University, Wuhan, China

[2]School of Remote Sensing and Information Engineering, Wuhan University, Wuhan, China

*E-mail address:* zhengxw@whu.edu.cn; whu_hlx@whu.edu.cn

June 7, 2019

## Abstract

Semantic segmentation has been a hot topic across diverse research fields. Along with the success of deep convolutional neural networks, semantic segmentation has made great achievements and improvements, in terms of both urban scene parsing and indoor semantic segmentation. However, most of the state-of-the-art models are still faced with a challenge in discriminative feature learning, which limits the ability of a model to detect multi-scale objects and to guarantee semantic consistency inside one object (i.e. intra-class consistency) or distinguish different adjacent objects with similar appearance (i.e. inter-class distinction). In this paper, a practical and efficient edge-aware neural network—ELKPPNet—is presented for semantic segmentation. This end-to-end trainable engine consists of a new encoder-decoder network, a large kernel spatial pyramid pooling (LKPP) block, and an edge-aware loss function. The encoder-decoder network was designed as a balanced structure to narrow the semantic and resolution gaps in multi-level feature aggregation, while the LKPP block was constructed with a densely expanding receptive field for multi-scale feature extraction and fusion. Furthermore, the new powerful edge-aware loss function is proposed to refine the boundaries directly from the semantic segmentation prediction for more robust and discriminative features. The effectiveness of the proposed model was demonstrated using two popular outdoor datasets—Cityscapes and CamVid—and the NYUDv2 indoor benchmark dataset. The performance of the two structures and the edge-aware loss function in ELKPPNet was validated on the Cityscapes dataset, while the complete ELKPPNet was evaluated on the CamVid and NYUDv2 datasets. A comparative analysis with the state-of-the-art methods under the same conditions confirmed the superiority of the proposed algorithm. A TensorFlow implementation of ELKPPNet will be available at https://github.com/XianweiZheng104/ELKPPNet.

## 1 Introduction

Semantic segmentation is a dense classification task that assigns categorical labels to every pixel in an image, with a wide range of applications. These include autonomous driving, outdoor and indoor scene parsing (Li and Yang 2016; Jiang et al. 2017), urban 3D semantic modeling (Rouhani et al. 2017; Zhang et al. 2018), support relationship inferences (Yang et al. 2017) and remote sensing image classification (Kemker et al. 2018; Politz and Sester 2018). Semantic segmentation developed slowly because the various objects found in an image limit the efficiency of automatic scene parsing, until the popularization of deep learning. Since then, deep convolutional neural networks (DCNNs) have enabled semantic segmentation algorithms to achieve remarkable progress in RGB scenes. As DCNNs have the ability to learn rich hierarchical features, they have become the mainstream approach for scene understanding.

Since the success of the first end-to-end deep neural network for semantic segmentation (Long et al. 2015), much effort has been devoted to the improvement of semantic segmentation accuracy. Despite the impressive results achieved by recent networks in some scenarios, there are still two main challenges. The first problem exists in multi-scale object detection. In semantic segmentation, when there is scale variance in an image, a neural network with an inappropriate receptive field size will give unbalanced attention to multi-scale objects. A neural network with small view field will pay more attention to small things and divide the larger objects into several smaller parts, while one with large view field will ignore details and fail to separate small adjacent objects. In mainstream semantic segmentation architectures, common methods focus on receptive field diversity and enlargement. As shown in Fig. 1, these methods can be grouped into four categories: 1) methods based on image pyramids (Zhao et al. 2018); 2) methods using an extra subnetwork (Yang et al. 2018); 3) methods applying an encoder-decoder structure (Badrinarayanan et al. 2017); and 4) methods deploying spatial pyramid pooling (SPP) (Zhao et al. 2017; Chen et al. 2017). However, the image pyramid and extra subnetwork approaches involve a high computational cost, while the encoder-decoder methods encounter semantic and resolution gaps in the multi-level features. The current SPP modules either have a large number of parameters or lack the necessary relational information between local features. Thus, multi-scale object detection remains a significant impediment to further improvements in segmentation accuracy.

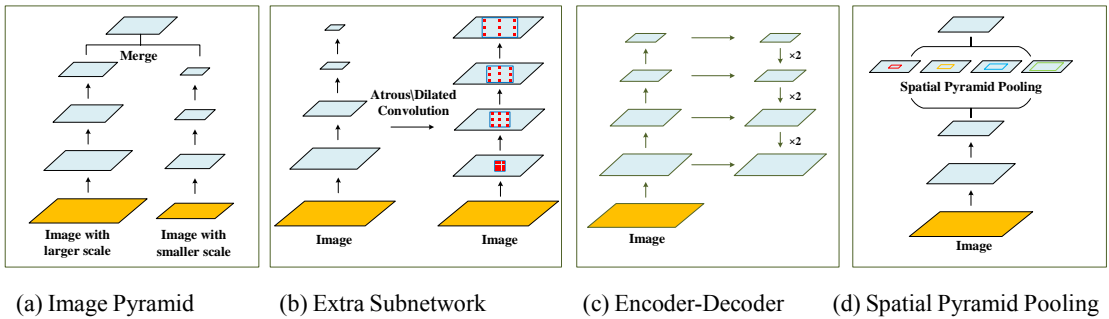

(a) Image Pyramid    (b) Extra Subnetwork    (c) Encoder-Decoder    (d) Spatial Pyramid Pooling

Fig 1. Existing methods for multi-scale context extraction (Chen et al. 2017).

The other difficulty for precise semantic segmentation lies in detail refinement. Most deep learning methods are not sensitive to detail information, and thus they often cannot maintain semantic consistency inside a single object (intra-class inconsistency) or distinguish two semantically different adjacent objects with similar appearances (inter-class indistinction) (Yu et al. 2018). In both instances, the boundaries between objects will have ambiguous contours that might confuse the network. For example, if a line between shadow and light on a desk is viewed as part of an object contour, then a model may be misguided and divide the desk into two parts; similarly, if the boundary between two semantically different objects is ignored, a model will recognize them as one object. A general solution for these two issues is to strengthen the ability of a model to extract edge information, which can determine the localization of an object. However, most of the current methods rely on post-processing techniques or an extra subnetwork for edge information capture, both of which fail to refine the contours in semantic segmentation prediction directly, resulting in expensive computational requirements.

Based on these observations, we have reformulated these challenges as a discriminative feature learning problem, and proposed an edge-aware deep neural network, termed as ELKPPNet, for robust pixel-wise semantic segmentation. The proposed ELKPPNet is composed of two newly developed structures, i.e., a balanced encoder-decoder and a large kernel pyramid pooling (LKPP) module, which are combined with a novel powerful edge-aware loss function to learn discriminative features. In ELKPPNet, the balanced bottom-up encoder-decoder narrows the semantic and spatial gaps in the multi-level features with simple transformation layers, which are more resourceful than the sophisticated blocks used by Lin et al. (2017) and Zhou et al. (2018). This approach also refines the semantic boundary for prediction using geometrical information from the low-level features. The SPP module, i.e., LKPP, is constructed with large kernels with hybrid asymmetric dilated convolutions to overcome the limitations of the existing SPP modules. The LKPP module can encode rich spatial information at a low computational cost, thus alleviating the "gridding" problem caused by dilated convolution (Wang et al. 2018). Furthermore, to address the "intra-class inconsistency" and "inter-class indistinction" issues, an edge-aware loss function is introduced to directly refine the semantic edge details and model the semantic dependency of regions.

The rest of this paper is organized as follows. The related work is reviewed in Section 2. The framework of ELKPNet and its components are detailed in Section3. The performance of the two new structures, the novel edge-aware loss function and the complete ELKPNet is evaluated in Section 4. Finally, our conclusions are drawn in Section 5.

## 2 Related work

### 2.1 Multi-scale Object Detection

Scale variance of objects occurs frequently in natural and remote sensing images, and influences the learning ability of deep networks for semantic segmentation. Many researches have focused on enhancing the robustness to scale variance by view field enlargement and effective multi-level feature fusion. The most powerful approaches are the image pyramid technique, the extra sub network method, the encoder-decoder structure, and the SPP architecture.

Although these methods have been widely used, they still have drawbacks and limitations. An image pyramid works by parallel networks to learn features from inputs of several sizes, where the long-range context and small object details become prominent with different input sizes. Typical examples of such methods include the method developed by Eigen and Fergus (2016) , which uses coarse-to-fine multi-scale inputs, and the methods proposed by Farabet et al. (2013), Zhao et al. (2018), and Chen et al. (2016), which feed images of multiple scales to a network and then merge the feature maps from all the scales. The main shortcomings of the image pyramid approach are that it usually requires a lot of graphics processing unit (GPU) memory, a long training time for multi-size inputs, and prior knowledge or extra convolutional layers for multi-size output fusion. Furthermore, it does not combine multi-level information from parallel networks.

The extra subnetwork approach performs multi-scale feature detection by cascading a feature extractor on top of the main network. Some methods add DenseCRF (Krähenbühl and Koltun 2011) to model region similarities (Zheng 2015; Li 2016; Chen 2016), and some adopted several sequential convolutional layers to extract long-range information (Yu and Koltun 2016; Liu et al. 2015). DenseASPP involves organizing atrous convolutional layers with increasing rates in a dense fashion to enlarge receptive filed size (Yang et al. 2018). However, in practice, the extra subnetwork brings heavy computational complexity and a high memory footprint. Structures such as DenseASPP, in particular, suffer from high computational cost and optimization trouble coming from the dense connection and concatenation.

The encoder-decoder framework achieves multi-level feature aggregation by merging low-level and high-level features. Due to the fact that low-level features are rich in spatial details but lack semantic information, while high-level features are just the opposite, semantic and resolution gaps exist between the multi-level features. This means that it can be inappropriate to directly fuse low-level high-resolution and high-level low-resolution features, as in U-Net (Ronneberger et al. 2015). To alleviate the semantic and resolution gaps in the multi-level features, RefineNet introduces a complicated refinement module between the encoder and the decoder (Lin et al. 2017), instead of a simple skip connection. U-Net++ employs a deeply supervised encoder-decoder network (Zhou et al. 2018), where the encoder and decoder subnetworks are connected through a series of nested dense skip pathways. ExFuse performs feature aggregation by deeply supervised learning (Zhang et al. 2018). Jegou et al. (2016) and Li et al. (2019) constructed dense multi-scale connections for feature aggregation, and Yu et al. (2018) hierarchically fused multi-level features by deep layer aggregation. However, these methods often need well-designed aggregation structures, which require prior knowledge and introduce a large number of parameters, which come with a high computational cost.

Spatial pyramid pooling (SPP) captures rich context features by multiple effective receptive fields, which come from parallel pooling operations or convolution layers. Since the successful application of spatial pyramid pooling in object detection task (He et al. 2014), many applicable variants have been put forward, specifically for semantic segmentation. For example, the pyramid pooling module in PSPNet is composed of convolutions with kernels of four varying sizes (Zhao et al. 2017), while atrous spatial pyramid pooling (ASPP) module proposed in DeepLabv2 groups parallel atrous convolution layers with different rates (Chen et al. 2016). However, the drawback of these current popular SPP structures lies in three aspects: 1) the SPP with standard convolution has difficulty in expanding the receptive field by a large kernel size, since increasing the kernel size will result in a large number of parameters; 2) the SPP with small kernels (e.g. ASPP) lacks enough connection between local features; and 3) the "gridding" problem, which happens when the view field is enlarged by dilated convolutional layers (Wang et al. 2018).

In the proposed network, the balanced encoder-decoder framework is capable of efficient and computation-saving multi-level feature aggregation, and the novel spatial pyramid pooling module—LKPP—can obtain highly rich contextual features with a relatively low computational burden and no missing information.

**2.2 Discriminative Feature Learning**

Deep neural networks are famous for their ability to learn robust and discriminative feature representations. However, in situations where objects are adjacent or have a similar appearance, the features extracted through a neural network are often poor at separating neighboring items or maintaining semantic consistency. This is because neural networks tend to learn local texture features instead of a geometric prior, and over-smooth the edges of objects, which include significant information for localization. To capture the local relationship of pixels and contour details, most models utilize post-processing techniques such as conditional random fields (CRFs) or edge maps yielded by a subnetwork. Popular models such as those developed by Li and Yang (2016) and Chen et al. (2016) capture fine details by employing DenseCRF (Krähenbühl and Koltun 2011), while Yu et al. (2018) combined semantic segmentation and boundary detection by two subnetworks—Smooth Network and Border Network—to address the "intra-class inconsistency" issue and enlarge the "inter-class distinction". Jiang et al. (2017),

Lee et al. (2017) and Marmanis et al. (2018) extracted edge features from DEM data or a depth map. However, techniques such as DenseCRF are often computationally expensive, and extra data such as a DEM or depth map are not always available. Furthermore, the edge maps predicted with networks presented by Xie et al. (2015) and Liu et al. (2017) are usually over-smoothed, which results in local ambiguities for accurate segmentation, and they may even distort object outlines.

In fact, the segmentation labels and predictions contain rich and fine contour details, but most works just ignore such advantageous edge information and fail to directly optimize the edges from the semantic segmentation prediction. In this study, an edge-aware loss function was designed to take advantage of this direct edge information source, and force the network to concentrate more on discriminative features for precise boundary detection, which plays a key role in achieving high intra-class consistency and clear-cut inter-class distinction.

## 3 Architecture

This section first describes the overall workflow of the proposed ELKPPNet, and then gives a detailed elaboration of the working mechanism of the balanced encoder-decoder structure, the main components of ELKPPNet (including the construction of the LKPP module and its related theories), and the design of the edge-aware loss function.

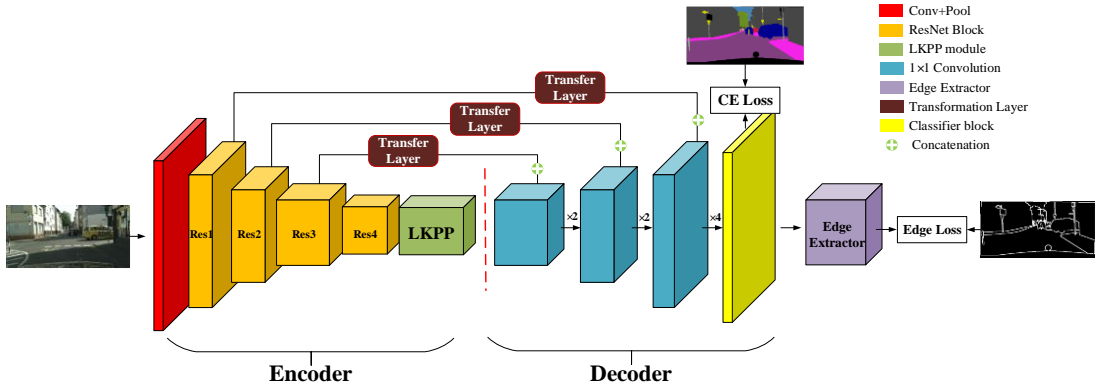

Fig. 2. The whole network architecture of the proposed ELKPPNet.

### 3.1 The Workflow of ELKPPNet

The proposed ELKPPNet framework, as illustrated in Fig. 2, features a residual network as an encoder, and a decoder followed by a classifier layer and an edge extractor. ELKPPNet takes an RGB image as input, and outputs a semantic segmentation prediction at the classifier layer with an edge map through the edge extractor, which directly processes the segmentation result. In the whole framework, the four residual blocks (Res1, Res2, Res3, and Res4) first encode the input image, and the LKPP module uses the features in the fourth block for richer contextual feature learning. During encoding, the image increasingly losses its representation details, which is not beneficial for the boundary delineation in segmentation. It is therefore necessary for the decoder network to reuse the boundary information stored in the low-level encoder feature maps. In order to recover accurate object structures, the decoder part upsamples the output of the LKPP module with the multi-level encoder features delivered by the transfer layers, which reduces the semantic and resolution gaps in the multi-level feature aggregation and adds long-range contextual information into the decoding path. Such a framework with transfer layers helps keep a balance between semantic information extraction and geometric structure protection, and is thus called the balanced encoder-decoder architecture. At the end of the decoding, the semantic segmentation map separately goes through a softmax activation and a sigmoid function for probability prediction. The softmax probability map is used to form the standard cross-entropy loss, and the sigmoid one is processed by the edge extractor to infer the object contours, which are modified by the edge loss function for direct semantic detail refinement and discriminative feature learning. The final combination of standard cross-entropy loss and edge loss is named edge-aware loss for its detail mining function. In the training stage, the whole of ELKPPNet is trained by this edge-aware loss function.

### 3.2 Feature aggregation with Balanced Encoder-Decoder

With the ability of aggregating multi-level features, the encoder-decoder network has become a popular framework in generic networks. A standard encoder-decoder neural network has two components: an encoder path and a decoder part. In many frameworks, these components are connected through simple or well-designed skip pathways. With the skip connections, a network can recover fine-grained details of the target objects and generate detailed segmentation masks, even against a complex background. U-Net adopts direct concatenation between the same-level layers in the encoder and decoder (Ronneberger et al. 2015), while other networks such as U-Net++ (Zhou et al. 2018) and RefineNet (Lin et al. 2017) choose complex structures.

However, in multi-level feature aggregation, there is an important issue that should be noted: the semantic and resolution gaps between layers. Theoretically, straight concatenation (as in U-Net) may fail to finely fuse the features from the layers at several levels in the decoder, and may propagate unnecessary semantic information to the lower layers. Such a characteristic may even introduce information confusion into the whole network, and impede low-level feature learning in the bottom layers during training. Focusing on this issue, the balanced encoder-decoder employs a simple 1×1 convolution as the transfer layer in each skip pathway, to narrow the semantic and resolution gaps. Unlike the complex well-designed skip pathways in models such as U-Net++ and RefineNet, the transfer layer acts more like a buffer zone. In backpropagation, gradients are first calculated in the transfer layer, and this prevents direct parameter update in the lower layers of the encoder subnetwork. The transfer layer can thus avoid having a serious impact on the low-level feature learning. Moreover, after the transformation of the transfer layer, features from multiple levels will share similar semantic information while preserving discriminative details. This makes it easier for the proposed model to more effectively capture fine-grained details of the objects. See Fig. 2 for an illustration of the whole encoder-decoder network framework.

### 3.3 Multi-scale Feature Capture with Large Kernel Pyramid

SPP module deployed in atrous (dilated) convolution fashion has been proved to be able to effectively extract denser multi-scale high-level feature maps, without downsampling operations or extra parameter learning. To learn dense contextual features without "gridding" effect brought by dilated convolutions, a new SPP module based on hybrid asymmetric dilated convolution, namely LKPP, is introduced in this part.

### 3.3.1 Atrous (Dilated) Convolution

Atrous convolution, which is also known as dilated convolution, is equivalent to inserting holes (zeros) into kernel weights. For a feature map $X$ and its corresponding output $Y$, a 2-D atrous convolution with filter $w$ is defined as

$$Y[i] = \sum_{k} X[i + r \cdot k] \cdot w[k] \qquad (1)$$

where $i$ and $k$ denote the location in the output $Y$ and filter $w$, respectively. The atrous rate $r$ corresponds to the stride. For a dilated convolution with a filter of size of $k \times k$, its dilated filter covers a region of size $k_d \times k_d$, where $k_d = k + (k-1) \cdot (r-1)$, as illustrated in Fig. 3(a). By changing the rate of an atrous convolution, the receptive field size of its filters can be effectively enlarged, without pooling or strided convolution.

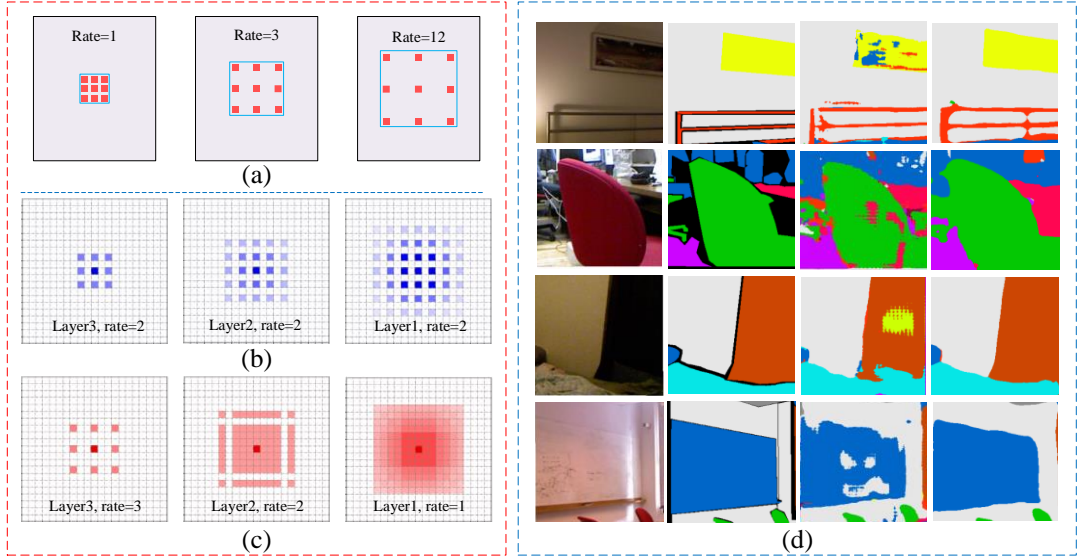

Fig. 3. Gridding problem of Atrous Convolution. (a) An example of atrous convolution (Chen et al. 2017). (b) Subsequent convolutional layers with the same dilation of 2. (c) Subsequent convolutional layers with dilation rates 1, 2, 3. (d) Effectiveness of LKPP module in eliminating the gridding effect: the first column is the original image patch, the second column is the ground truth patch, the third column indicates the gridding effect result from Atrous Convolution, and the fourth column is the prediction result of ResNet-50 (He et al. 2016) with the proposed LKPP module. It can be seen that gridding effect produces checkboard-like patterns, while the proposed LKPP effectively eliminate such an effect.

As demonstrated in Chen et al. (2017), the larger the dilation rate grows, the small the number of effective kernel weights will become. For example, if the filter size is close to the feature map size, it will simply work like a 1×1 filter. To expand the view field without degeneration, it is advisable to use stacked atrous convolution layers with large rates, as in DenseASPP (Yang et al. 2018), but such a solution also causes the problem known as "gridding" (Wang et al. 2018). Taking $k = 3$ and $r = 2$ for illustration, if a group of sequential convolutional layers have the same rate $r$, then given an arbitrary pixel $p$ of the top layer $l_i$, its receptive field is formed in a checkerboard fashion, meaning much of the information from the input is discarded, as indicated in Fig. 4(b). Such an issue may cause three obstacles for dense prediction: 1) the local information may be impaired; 2) the long-distance information may be irrelevant; and 3) the consistency of the local features may be severely violated, since the information of pixels in the neighboring $r \times r$ regions of $l_i$ may be contributed by uncorrelated pixels in previous layer $l_{i-1}$.

### 3.3.2 Hybrid asymmetric dilated convolution

Hybrid dilated convolution (HDC) (Wang et al. 2018) is a solution to address the 'gridding' issue.

Given $N$ convolutional layers $\{l_1,...,l_N\}$ with kernels of size $k \times k$ chained in cascade, and $\{r_1,...,r_N\}$ denote their dilation rates, we can define the maximum distance between nonzero weights at layer $l_i$ as:

$$M_i = \max(|M_{i+1} - 2r_i|, r_i) \tag{2}$$

where $M_N = r_N$. If $M_2 \leq k$, the final size of the receptive field of the convolution series $\{l_1,...,l_N\}$ will fully cover a large square region, without any holes or missing edges.

Although original HDC structure can view a large region without information loss through sequent layers, the features from one layer will still not be richer enough due to the fixed small kernel size, which can lead to a lack of discriminative features. Actually, in a series of layers $\{l_1,...,l_N\}$, layer $l_i$ obtains higher-level features by mixing features from the lower layer $l_{i-1}$, which indicates that richer information in layer $l_{i-1}$ could be helpful for the feature extraction in layer $l_i$. Richer information is usually obtained by large kernel, but it is hard to expand the kernel size with standard convolution since the kernel size grows in the form of $k^2$, which leads to high calculation cost.

Inspired by Szegedy et al. (2016), we factorize the dilated convolution of HDC with asymmetric filters of $k_1 \times k_2$ size, where $k_1 \neq k_2$ and $\max(k_1, k_2) > 3$, generating filters of rectangle shape. Zeros are inserted along the dimension with $\max(k_1, k_2)$, while the dilation rate in the other dimension maintains $1$. Consider, for example, the case of a $5 \times 1$ convolution followed by another $1 \times 5$ convolution, the receptive field of this two-layer block is the same as that of a $5 \times 5$ convolutional layer, while the number of parameters is 60% smaller. Unlike the $1 \times k$ or $k \times 1$ filters used in Szegedy et al. (2016), which easily ignores the local relationship of a certain dimension, our asymmetric dilated convolution satisfies the following condition, to encode local relationship information of both dimensions:

$$\min(k_1, k_2) \geq 2 \tag{3}$$

In this case, the hybrid asymmetric dilated convolution (HADC) block, formed by three two-layer modules, can extract rich contextual information with large kernels and low computational burden, and enlarge the whole receptive field without the "gridding" problem.

### 3.3.3 Large Kernel Pyramid Pooling

The proposed LKPP module is a SPP module based on HADC. In the proposed method, LKPP includes three HADC blocks to detect objects at different scales, and fuses local and long-range features. The two-layer convolution in an HADC block can be arranged in two ways for semantic segmentation in different situations—in cascade fashion or parallel fashion—as illustrated in Fig. 4.

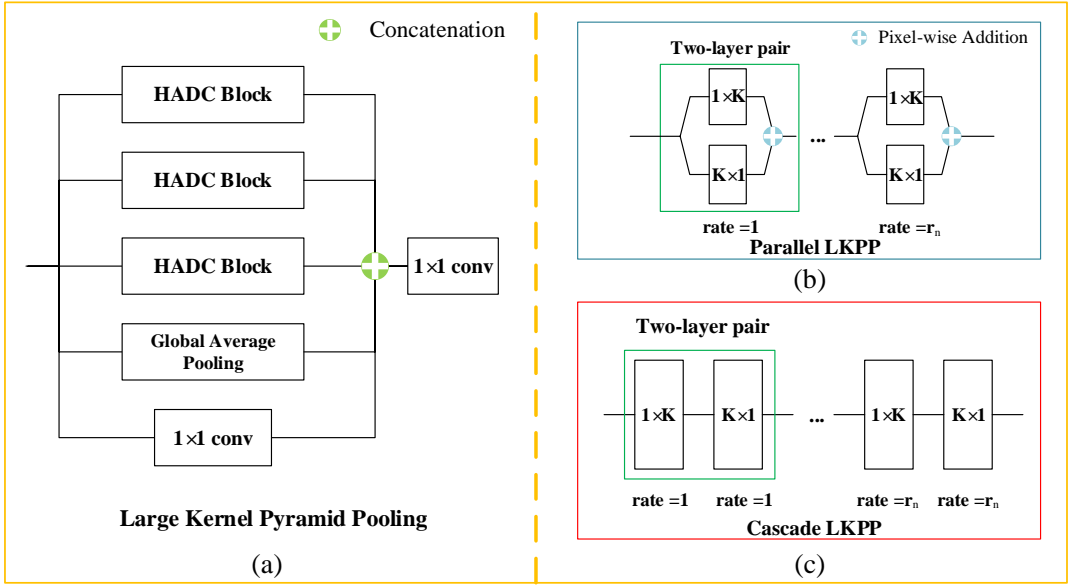

Fig. 4. The two-layer convolution in an HADC block. (a) Large Kernel Pyramid Pooling. (b) The HADC block in parallel LKPP. (c) The HADC block in cascade LKPP.

**Cascade LKPP:** In cascade LKPP, each HADC branch consists of three two-layer pairs, and the layers in each pair are joined sequentially, which can greatly expand receptive field size, and therefore is applicable to outdoor scenes, where many large-scale and regular objects exist, such as buildings, streets and cars.

**Parallel LKPP:** In parallel LKPP, the two convolutions in each two-layer pair are placed in parallel and merged by pixel-wise addition, in which case features in two dimensions will be equally treated. Unlike cascade LKPP that sees a rectangle area, parallel LKPP extracts cross-shaped features. Compared to the cascade structure, receptive field in parallel LKPP grows more slowly so as to capture features in a denser way, and its irregular shape tends to find objects of unfixed shape, e.g., sheets and clothes. Therefore, the parallel LKPP is more suitable for indoor scenes, since indoor scenes are generally characterized by occlusion and overlap caused by limited and enclosed space.

In LKPP, we also apply a 1×1 convolution as a skip connection to reuse the original signals. To further alleviate the "gridding" effect and incorporate global contextual information, we compute image-level features by a global context branch made up of a global average pooling operation followed by a 1×1 convolution and batch normalization (Ioffe and Szegedy 2015). The image-level features are then reshaped into the desired spatial dimension by bilinear interpolation, for later concatenation with the features from other branches.

### 3.4 Detail refinement with Edge-aware Cross-Entropy loss

Softmax with cross-entropy loss, which is also referred as cross-entropy loss (CE loss), is the objective function most often used in the semantic segmentation task. It has many varieties, e.g., weighted softmax loss (Xie and Tu 2015), focal loss (Lin et al. 2017), etc., to help excavate fine details. In a DCNN-based semantic segmentation algorithm, over-smoothing is a common problem that affects the segmentation quality, since it can impair the boundary between two adjacent objects, making it difficult to distinguish apparently similar but semantically different objects or keep semantic consistency within an object. For example, holistically nested edge detection often produces over-smoothed border

(Xie and Tu 2015), and DeepLab tends to yield droplet-like patches (Chen et al. 2017), which may fill the gap between two separate parts. In this part, we focus on directly refining the edge map of the semantic segmentation results by a novel edge-aware cross-entropy loss (ECE loss) function.

### 3.4.1 Cross-entropy loss

Given the feature map of the classifier layer as $Z \in \mathbb{R}^{H \times W \times C}$ (where $H, W, C$ represent the height, width and channel of the feature map, respectively, and $C$ is equal to the number of classes), the prediction map produced by softmax layer is defined as (4):

$$p(y_{ij} = k | I, \Theta) = e^{Z_{ijk}} / (\sum_{c=1}^{C} e^{Z_{ijc}}) \tag{4}$$

where $I, \Theta, y$ denote the input image, the model parameters, and the predicted label, respectively, and $p(y_{ij} = k | I, \Theta)$ refers to the probability of a pixel at location $(i, j)$ belonging to the object denoted by class $k$. If we let $\hat{y}, \hat{p}$ be the ground truth and the corresponding probability, then the segmentation loss takes the form of (5).

$$l_{seg} = -\sum_{i,j,k} \hat{p}(\hat{y}_{ij} = k | I) \ln p(y_{ij} = k | I, \Theta) \tag{5}$$

Although cross-entropy loss is adopted in many methods, it does not directly consider the relationship between different objects. Actually, it treats every pixel separately in the semantic map without linking pixels with their neighbors, which is where the proposed edge-aware loss works.

### 3.4.2 Edge Loss

In image processing, edge detection is often performed by different gradient-based and mathematical morphological techniques. Defined by the divergence of the gradient, Laplacian operator (Gonzalez and Wintz 2002) is a simple second order differential operator, and its 2-D form is defined by (6).

$$\Delta f = \frac{\partial^2 f}{\partial x^2} + \frac{\partial^2 f}{\partial y^2} \tag{6}$$

In digital image processing, a 2-D Laplacian operator consists of a $3 \times 3$ mask and has an extended template (7).

$$\begin{bmatrix} 0 & 1 & 0 \\ 1 & -4 & 1 \\ 0 & 1 & 0 \end{bmatrix} \quad \begin{bmatrix} 1 & 1 & 1 \\ 1 & -8 & 1 \\ 1 & 1 & 1 \end{bmatrix} \tag{7}$$

Because the segmentation map contains rich edge information and Laplacian operator is suitable for signal without noise, where it could yield strong responses at boundary, we thus choose Laplacian operator to directly draw gradient information from the semantic segmentation prediction map. The output of the Laplacian operator is defined as gradient map. Here, we define a more general Laplacian operator template with flexible receptive field (8):

$$L_k = \begin{bmatrix} I_{k \times k} & I_{k \times k} & I_{k \times k} \\ I_{k \times k} & -8I_{k \times k} & I_{k \times k} \\ I_{k \times k} & I_{k \times k} & I_{k \times k} \end{bmatrix} \tag{8}$$

where $I_{k\times k}$ is a $k\times k$ identity matrix, and the Laplacian operator is a special case when $k=1$. The value of $k$ determines how much information should be considered in the gradient computation. The bigger $k$ is, the thicker detected edge line is, i.e., pixels close to edge will have strong responses, as illustrated in Fig. 5(a).

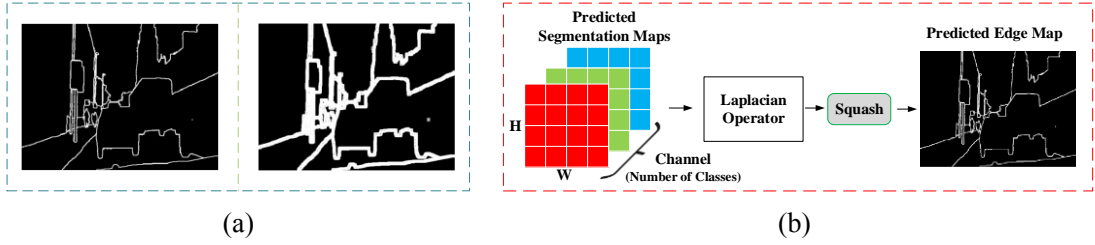

(a)                                             (b)

Fig 5. Edge Extractor. (a) Edge map under different $k$. Left: Edge map with $k=1$; Right: Edge map with $k=3$. (b) The mechanism of edge extractor.

Edge detection is a binary classification problem, but the gradient map only contains semantic edge information and optimizing semantic edge, may introduce unnecessary computation and require more GPU memory, as in Liu et al. (2018). In this case, a function is needed to convert the gradient map into an edge probability map.

Given a gradient map $G \in \mathbb{R}^{H\times W\times C}$, the $1\times 1\times C$ vector at every position $(i,j)$ records the gradient information for every class. To obtain the edge probability map, a squash function (9) is further applied to calculate the gradient intensity, and to compress the intensity value into $[0,1]$, to represent the probability of a pixel belonging to an edge, as shown in Fig. 5(b).

$$e_{ij} = squash(\vec{g}_{ij}) = \frac{\|\vec{g}_{ij}\|}{\|\vec{g}_{ij}\| + \alpha} \tag{9}$$

In (9), $\vec{g}_{ij}$ and $e_{ij}$ respectively denote the vector and edge probability at $(i,j)$, $\|\cdot\|$ is the $l_2$-norm, and $\alpha$ is a hyper-parameter that controls the model's sensitivity to "intra-class inconsistency" and "inter-class indistinction". In fact, if $\|\vec{g}_{ij}\| > \alpha$, the edge probability $e_{ij}$ at location $(i,j)$ will be greater than 0.5, meaning that the pixel at $(i,j)$ is more likely to belong to an edge. Intuitively, when $\alpha$ is small, pixels with a weak response (i.e. small $\|\vec{g}_{ij}\|$) will be viewed as part of the edges, and slight oscillation will result in the wrong edge inside an object. With a small value of $\alpha$, the model pays more attention to correcting the wrong edge predictions inside objects, and therefore enhances the intra-class consistency; similarly, when $\alpha$ is large, the model focuses more on modifying the boundary between two adjacent objects, and instead magnifies the inter-class distinction. In our experiments, $\alpha$ was set to 1 to make a balance between inter-class distinction and intra-class consistency.

Given the predicted edge map $E = [e_{ij}]_{H\times W}$ and the label $\hat{E} = [\hat{e}_{ij}]_{H\times W}$, the edge loss is formulated as a weighted binary cross-entropy loss as in (10):

$$l_{edge} = -\sum_{i,j}\left(\gamma\beta\hat{e}_{ij}\ln e_{ij} + (1-\beta)(1-\hat{e}_{ij})\ln(1-e_{ij})\right)$$
$$\beta = \frac{|\hat{E}^-|}{|\hat{E}^+| + |\hat{E}^-|} \tag{10}$$

where $|\hat{E}^+|$ and $|\hat{E}^-|$ denote the number of positive and negative samples, respectively. The hyper-parameter $\gamma$ is used to balance the number of edge and non-edge samples. In our experiments, we chose

$\gamma=1$.

### 3.4.3 Edge-aware cross-entropy loss

In training the whole semantic segmentation model, the objective function is defined as a combination of standard cross-entropy and edge loss as in (11):

$$ECELoss = l_{seg} + \lambda_1 l_{edge} + \lambda_2 \|\Theta\|_2 \tag{11}$$

In (11), $\lambda_1$ is a penalty parameter for balancing the standard cross-entropy term and the edge fidelity term, $\lambda_2$ is the decay weight, and $\|\Theta\|_2$ is the $l_2$-norm of all the model parameters. In ECE loss, the penalty parameter $\lambda_1$ controls the strength of the edge fidelity term. When $\lambda_1=0$, ECE loss is equivalent to the standard CE loss (cross-entropy loss). As $\lambda_1$ increases, the edge fidelity term gets more importance, so that the detail refinement gets more attention, and the network will be more sensitive to the gradient changes in the predicted segmentation map. Therefore, an appropriate $\lambda_1$ can assist the model in learning features more related to the object contours, but a large $\lambda_1$ may make the network more sensitive to the possible edge texture in the original images, which can be inconsistent with the edge ground truth (e.g. the grid pattern on the window). In practice, we set $\lambda_1$ as 0.5 to prevent learning features that are too distinctive to weaken the generalization ability, and we add the commonly used regularization term, $\|\Theta\|_2$, to prevent over-fitting.

## 4 Experiments

For the purpose of evaluation and comparison, the ResNet-50 (He et al. 2016) network was selected as the backbone for all the models, and the experiments were conducted on three challenging semantic segmentation datasets: the Cityscapes (Cordts et al. 2015; Cordts et al. 2016), and CamVid (Fauqueur et al. 2007; Badrinarayanan et al. 2017) outdoor datasets and the NYUDv2 indoor scene parsing benchmark dataset (Silberman et al. 2012; Lee et al. 2017). Ablation studies were first conducted on the Cityscapes dataset to validate the proposed balanced encoder-decoder structure, the LKPP module, and the ECE loss function, respectively. To allow a comprehensive evaluation, the whole ELKPPNet was further compared with the state-of-the-art DCNN algorithms on the CamVid and NYUDv2 datasets, including DeepLab V3+ (Chen et al. 2018) and PSPNet (Zhao et al. 2017). In all the experiments, except for mirror flip, no extra training tricks were used, especially those related to detail augmentation and multi-scale detection, because other training tricks add more random information, making it difficult to determine whether the discriminative feature learning ability is boosted by the given model structures.

### 4.1 Datasets and Implementation Details

**Cityscapes**: The Cityscapes dataset is a popular benchmark for urban scene understanding (Cordts et al. 2016). It contains 5,000 2048×1024 images with manually annotated pixel-level labels. There are 19 semantic classes in this dataset. Samples in this dataset are split into three parts: a training set of 2,979 images, a validation set of 500 images, and a test set of 1,525 images.

**CamVid dataset**: The Cambridge-driving Labeled Video Database (CamVid) contains 32 semantic classes (with 0 referring to the void class, which is treated as the background) and 701 segmentation images (Fauqueur et al. 2007). The CamVid dataset is divided into training, validation, and test sets. In our experiments, there were 421 images in the training set, 112 images in the validation set, and 168 images in the test set.

**NYUDv2 dataset**: The NYU-Depthv2 dataset is a challenging indoor scene parsing dataset with 1449 densely labeled pairs of aligned RGB and depth images (Silberman et al. 2012). The comparison with the state-of-the-art methods was divided into two parts. The first task was to classify the objects into 13 indoor scene classes, and the second was to classify the objects into 37 classes. Since classes in indoor images are typically more varied than those in outdoor scenes, frequent partial occlusions are a big issue, making this one of the most challenging datasets. In our training and testing, only the RGB modality was used. The dataset was split into separate training and test sets: 795 training images and 654 test images for the 13-class dataset, and 1304 training images and 145 test images for the 37-class dataset.

## 4.2 Training Strategy

In the training, the background was masked out to alleviate the impact of inconsistent texture. All the models were trained by the Adam optimizer with $2.5\times10^{-4}$ as the initial learning rate, and a poly learning rate policy was employed where the initial learning rate was multiplied by $(1-\frac{iter}{total\_iter})^{0.9}$ after each iteration. The decay weight was set as $5\times10^{-4}$. We used TensorFlow (Abadi et al. 2016) to train and evaluate all the models.

In order to validate the proposed model's superiority in detail capture and discriminative feature learning, all the experiments were conducted under simple conditions with no data augmentation, especially random scaling and cropping, which is a key trick for most of the state-of-the-art models. Because random scaling and cropping actually acts like the image pyramid technique, i.e., different scaling rates randomly change the object size and the cropping helps the model focus on certain detailed discriminative features. With random scaling and cropping, it is difficult to determine the model's ability of multi-scale object detection and edge refinement. Therefore, in all studies, the images are only resized to a fixed proper shape for training.

## 4.3 Evaluation Metrics

The model prediction performance is comprehensively assessed by a group of metrics, where pixels labeled as void do not contribute to the score.

**(1) Pixel accuracy (PixelAcc) and mean pixel accuracy (MeanPixelAcc)**: Accuracy is a common metric that evaluates how frequently a prediction matches its corresponding label. Pixel accuracy evaluates the accuracy over the whole image, while mean pixel accuracy averages the accuracy across all categories, as defined in (13):

$$PixelAcc = \frac{TP}{TP+FP},$$
$$MeanPixelAcc = \frac{1}{N}\sum_{k=1}^{N}\frac{TP_k}{TP_k+FP_k} \tag{13}$$

where $TP$ and $FP$ are the numbers of true positive and false positive; $N$ is the number of classes, and $TP_k$ and $FP_k$ denote the $TP$ and $FP$ of class $k$.

**(2) Mean Intersection-over-Union (mIoU):** The mIoU stands out as a standard metric for segmentation purposes, due to its representativeness and simplicity. It averages the ratio between the intersection area and the union area across all classes, which reveals prediction quality of every class. mIoU is an indicator adopted in all existing semantic segmentation works, which is defined as follows:

$$mIoU = \frac{1}{N}\sum_{k=1}^{N}\frac{TP_k}{TP_k + FP_k + FN_k} \qquad (14)$$

where $TP_k, FP_k$ and $FN_k$ denote the true positives, false positives and false negatives of class $k$, respectively, determined over the whole test set.

Since the mIoU treats all classes with the same weight, small objects with a low occurrence frequency, which may be insignificant, can greatly influence the score. We also adopt frequency-weighted intersection over union (FWIoU) to compensate for the bias, which is defined as in (15):

$$FWIoU = \frac{1}{\sum_{k=1}^{N}TP_k + FN_k}\sum_{k=1}^{N}\frac{(TP_k + FN_k) \cdot TP_k}{TP_k + FP_k + FN_k} \qquad . \qquad (15)$$

### 4.4 Ablation Studies With the Cityscapes Dataset

For ablation studies, the ResNet-50 network was selected as the backbone, and the models were trained on the Cityscapes dataset with a single NVIDIA GTX 1080 GPU. The training images were all resized to 224×448, which raises a challenge in small object learning. The batch size was 8, and all the models were updated for 150K iterations. We examined the performance with different settings of the network by the best model (measured by mIoU) saved during the training.

### 4.4.1 Evaluation of the Balanced Encoder-Decoder Architecture

To validate the performance of the balanced encoder-decoder architecture, U-Net (Ronneberger et al. 2015) with ResNet-50 as the backbone was set as the baseline. The numbers of channels were respectively 512, 256, and 128 in the three stages of the decoder. From Table 1, we can see that the gradual feature aggregation greatly increases all the indicators, especially the mIoU, which is increased by 6%. As shown in Fig. 6, it is clear that the balanced encoder-decoder framework captures more accurate boundaries, while the U-Net structure yields droplet-like over-smoothed patches.

Table 1. Results of U-Net and the balanced encoder-decoder network with CELoss (unit: %) .

| Metric \ Model | mIoU | FWIoU | PixelAcc | MeanClassAcc |
|---|---|---|---|---|
| U-Net | 43.56 | 81.52 | 89.00 | 52.76 |
| Our Balanced encoder-decoder | 49.59 | 84.72 | 91.16 | 58.09 |

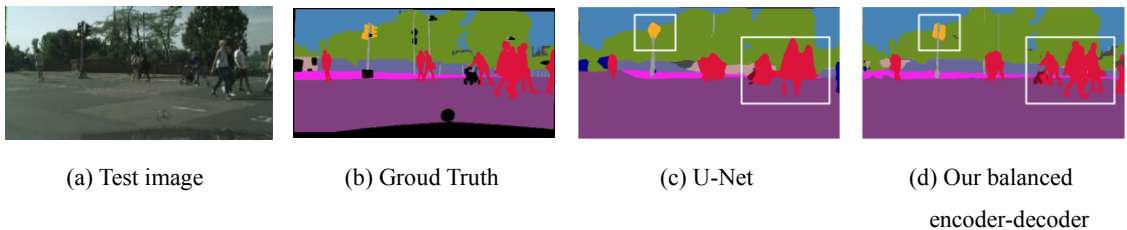

(a) Test image     (b) Groud Truth     (c) U-Net     (d) Our balanced encoder-decoder

Fig. 6. Prediction results of U-Net and the balanced encoder-decoder. In (c), the U-Net structure yields droplet-like

over-smoothed patches, which even erase the corners of the traffic sign (yellow) and distort its square shape into a nearly round one. In (d), the proposed balanced encoder-decoder framework more precisely draws out the contours of trees, pedestrians, and the street.

### 4.4.2 Evaluation of the LKPP Module

Since the 224×448 image is compressed into 7×14 in the encoder, where the spatial information is too heavily destroyed for any SPP module to learn multi-scale features, we added all the tested SPP modules on top of the network to extract multi-scale contextual information before the classifier block (a 3×3 convolution and a 1×1 convolution for classifying, as depicted in Fig. 3). Moreover, all the SPP modules were applied based on our balanced encoder-decoder architecture. The popular ASPP used in DeepLabv3+ (Chen et al. 2018), DenseASPP (Yang et al. 2018), and sequential atrous convolutions with the same dilation were chosen for comparison with the proposed LKPP module. Here, the network with stacked atrous convolutions is referred to DilatedNet, where the dilation rate was set as 2. In DilatedNet, the final upsampling operation helps to remove the visible "gridding" problem in the final output. The dilation rates of ASPP and DenseASPP were the same as Chen et al. (2018) (i.e., ASPP module with dilation rate of 1, 6, 12, 18) and Yang et al. (2018) (i.e., DenseASPP with dilation rate of 3, 6, 12, 18, 24). The kernels used in the LKPP module were set to 3×3, 3×5 (5×3), and 3×7 (7×3), and the rates in every HADC of the LKPP module were set as 1,2,3, to avoid superfluous invalid information caused by zero values introduced by large dilations. The baseline was a ResNet-50 network with a balanced encoder-decoder framework. All the models were updated by optimizing the standard CE loss.

From Table 2, by the mIoU metric, the LKPP module yields a 0.92% boost over ASPP, 3.76% over DilatedNet, 5.75% over DenseASPP, and 3.44% over the baseline. Different from mIoU, the metrics that are biased toward large objects (e.g. FIoU, PxielAcc) show little variance. Considering the fact that the mIoU treats objects equally at different scales, it can be concluded that the mIoU is boosted due to the successful small object or detail recognition, and thus the LKPP module is proved to be more sensitive to objects at a small scale, and is better at learning richer features. Moreover, the decrease in DilatedNet and DenseASPP demonstrates the loss of details caused by the "gridding" problem inside the networks.

Table 2. Quantitative results of several popular structures for multi-object detection (unit: %).

| Model \ Metric | mIoU | FWIoU | PixelAcc | MeanClassAcc |
|---|---|---|---|---|
| Baseline | 49.59 | 84.72 | 91.16 | 58.09 |
| ASPP | 52.11 | 85.61 | 91.78 | 60.45 |
| DilatedNet | 49.27 | 84.06 | 90.77 | 57.67 |
| DenseASPP | 47.28 | 84.17 | 90.84 | 55.31 |
| Our LKPP | 53.03 | 85.61 | 91.82 | 60.85 |

The qualitative results of the different structures for multi-object detection, as shown in Fig. 7 and Fig. 8, also demonstrate the effectiveness of the proposed LKPP structure in multi-scale feature capture. Compared to the results of the other models, as shown in Fig.7(c), (d), and (e), the proposed LKPP module draws out the large objects, e.g., the bus (in dark green) with correct annotation, while preserving the fine details of the small objects, e.g., the car (in blue) and the traffic sign (in yellow), as shown in Fig. 7(f). From the zoomed-in effect shown in Fig. 8, it can be seen that the bus window shares a similar texture with the tree and all the models fail to separate them, except the proposed LKPP-based model. This shows that the LKPP module excels at rich contextual information extraction and can learn more discriminative details for magnifying the inter-class distinction.

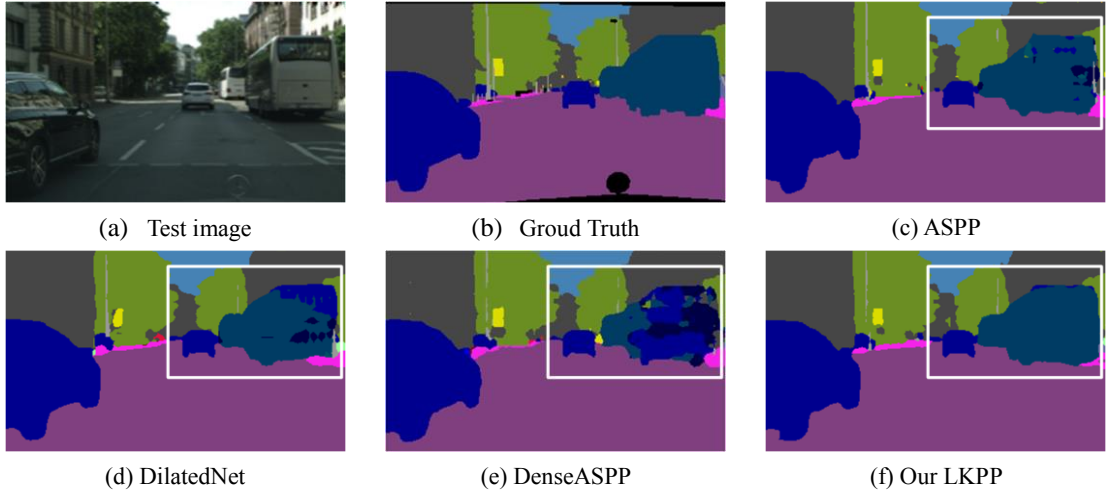

Fig. 7. Qualitative results of different atrous convolution based structures for multi-object detection.

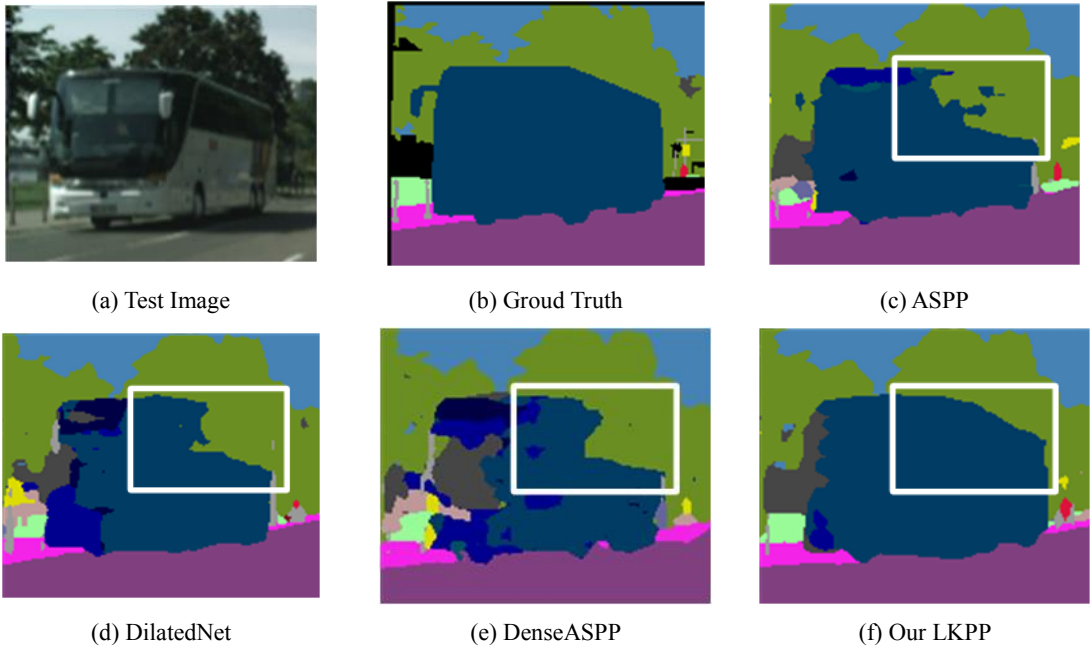

Fig. 8. Zoomed-in qualitative results.

### 4.4.3 Evaluation of Edge-aware Loss Function

A further evaluation was also made for the proposed edge-aware cross-entropy loss function (also referred as ECE loss). Resnet-50 with the proposed balanced encoder-decoder framework was applied as the baseline network, and the two loss functions, i.e., CE loss and the proposed ECE loss, were first studied on the baseline network. The two loss functions were then also used to train the proposed cascade LKPP network. From the results listed in Table 3, it is clear that the ECE loss function outperforms CE loss in all the metrics, on both the baseline network and the proposed cascade LKPP network, revealing the discriminative feature learning power of ECE loss.

Table 3 . Quantitative results of CE loss and ECE loss (unit: %).

| Metric / Model | mIoU | FWIoU | PixelAcc | MeanClassAcc |
|---|---|---|---|---|
| Baseline CE Loss | 49.59 | 84.72 | 91.16 | 58.09 |
| Baseline ECE Loss | 53.43 | 86.01 | 92.02 | 62.18 |
| LKPP CE Loss | 53.03 | 85.61 | 91.82 | 60.85 |
| LKPP ECE Loss | 54.68 | 86.50 | 92.36 | 63.20 |

Parts of the comparison results of ECE loss and CE loss are visualized in Fig. 8 and Fig. 9. Fig. 8 displays the effect of CE loss and ECE loss on the baseline network. It is noteworthy that the masked-out regions (the black regions in the ground truth) were excluded in the training, inference, and assessment, but there are still some unlabeled objects detected by the ECE loss, as shown in the rectangle highlighted in Fig. 8(d). Although this object has no annotation in the training set, the model trained by ECE loss still annotates the object with relatively good consistency, which could help a self-driving or navigation system to locate and avoid the untrained obstacles. In Fig. 8, when comparing (c) and (d), it can be seen that ECE loss is better at guiding the model to learn geometrical features to separate adjacent pedestrians (red) and carve out sharp contours for the traffic signs (yellow) and poles (gray). However, the features extracted by the baseline network are still not rich enough, even when using the ECE loss for training.

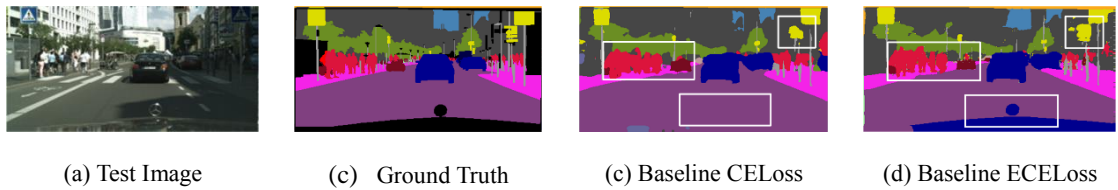

(a) Test Image  (c) Ground Truth  (c) Baseline CELoss  (d) Baseline ECELoss

Fig. 8. Qualitative results of CELoss and ECELoss on Baseline.

The results in Fig. 9 show that the ECE loss with the proposed LKPP network can learn more discriminative details. In Fig. 9(c), the LKPP network trained with CE loss can only segment different objects with coarse contours, and it still mistakes a small part of the bus (dark green) as a train (darker green). In Fig. 9(d), the LKPP network trained with ECE loss is capable of annotating the complete bus (dark green), and even draws out the bus wheels. Other details such as the pole (gray) and traffic sign (yellow) are also carved out with more accurate contours. All these demonstrate that the ECE loss can help to guarantee intra-class consistency and preserve more details.

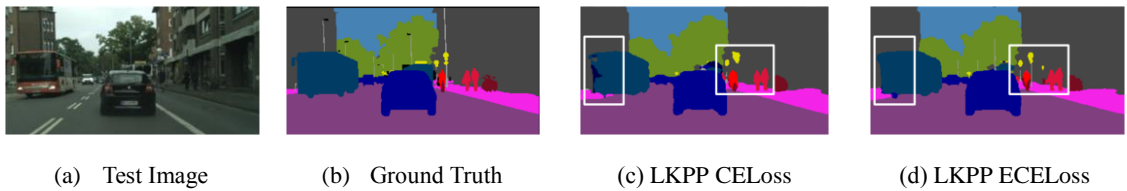

(a) Test Image  (b) Ground Truth  (c) LKPP CELoss  (d) LKPP ECELoss

Fig. 9. Qualitative results of CELoss and ECELoss on LKPP.

## 4.5 Results On the Camvid Dataset

For outdoor scene parsing, camvid dataset with 31 classes (excluding the void class) is chosen to evaluate the performance of the whole ELKPPNet, which composed of the complete balanced encoder-decoder, LKPP module, and ECE loss. Since there is a large imbalance in category frequencies, making it very challenging to label multi-scale objects. Thus it is particularly suitable to evaluate the effectiveness of the proposed ELKPPNet on discriminative feature learning. Similar to the ablation study on Cityscapes dataset, the training tricks except for mirror flip are discarded to validate different models' discriminative feature learning ability.

Table 4. Quantitative analysis on CamVid dataset (unit: %).

| Model \ Metric | mIoU | FWIoU | PixelAcc | MeanClassAcc |
|---|---|---|---|---|
| DeepLabV3 (Chen et al. 2017) | 40.41 | 80.24 | 88.54 | 44.25 |
| DeepLabV3+ (Chen et al. 2018) | 43.91 | 81.88 | 89.56 | 48.20 |
| DenseASPP (Yang et al. 2018) | 40.45 | 81.15 | 88.98 | 44.68 |
| PSPNet (Zhao et al. 2017) | 41.20 | 81.30 | 89.09 | 45.92 |
| RefineNet (Lin et al. 2017) | 40.46 | 81.38 | 89.21 | 44.13 |
| Ours ELKPPNet (cascade) | 46.45 | 82.94 | 90.18 | 54.07 |

All of the models used for comparison were built on the ResNet-50 network, due to the memory limitation, and were trained on a single NVIDIA Tesla V100 GPU with a batch size of 6 for 120K iterations. The images were resized to 384×512 for training and testing, which is compatible with the original resolution of 960×720. The quantitative results are listed in Table 4, where it can be seen that the proposed ELKPPNet exceeds all the other models in all metrics. Specifically, in Table 4, the mIoU and MeanClassAcc of ELKPPNet exceed the results of DeepLabv3+ by 2.54% and 5.87%, respectively. Since the FWIoU and PixelAcc are mainly contributed by large objects, the obvious improvement in mIoU and MeanPixelAcc indicates that the proposed ELKPPNet is more capable of detecting small objects and extracting details.

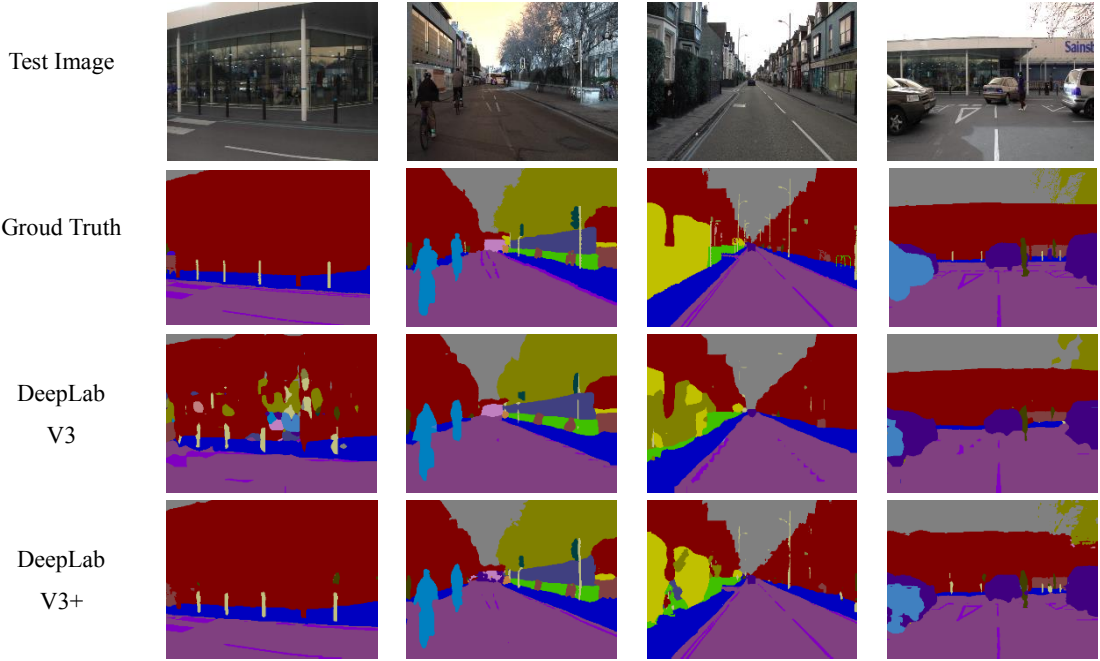

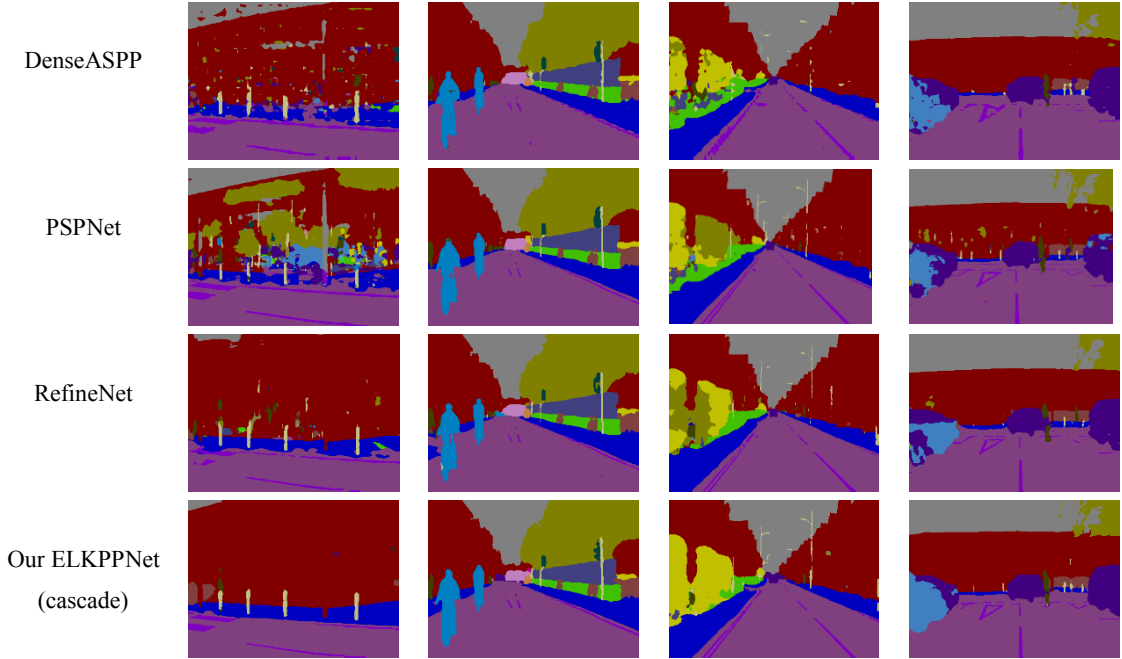

Fig. 10. Qualitative results on Camvid dataset.

In Fig. 10, ELKPPNet succeeds in both maintaining intra-class consistency with sharp edges and achieving inter-class distinction. It can be seen that ELKPPNet yields a cleaner prediction in the first column, meaning successful settlement of the intra-class inconsistency issue. Moreover, ELKPPNet obtains a more precise contour of the traffic light (dark green) and the truck bus (pink) in the second column, and better semantic consistency for the vegetation misc (yellow) in the fourth column. In the last column, in particular, ELKPPNet even successfully separates the pickup truck (sky blue) from the adjacent ordinary car (purple), which shares nearly the same appearance, showing the excellent ability of ELKPPNet to learn discriminative features for inter-class distinction.

### 4.6 Results On the NYUDv2 Dataset

For indoor scene understanding, NYUDv2 with 13 and 37 classes (objects, furniture, walls, ceilings etc., excluding the void class) were selected as the benchmark datasets, since the NYUDv2 dataset is a popular indoor parsing dataset for benchmarking multi-scale deep learning approaches. This benchmark is therefore useful to compare methods focusing on multi-scale issues. All the models were built on the ResNet-50 network and trained on a single NVIDIA Tesla V100 GPU. The results obtained with both datasets show that ELKPPNet is more suitable for indoor scenes than the other popular networks.

**13-class NYUDv2 dataset:** The 13-class dataset contains 795 training and 654 testing images, which is the separation used in most RGB-D based indoor segmentation methods. Since the number of classes is small, all the models were trained with a batch size of 4 for 100K iterations, and all the images retained the original resolution of 640×480. Because indoor scene understanding is sophisticated and it is uncertain as to which optimizer works best in such a situation, each model was trained with the Adam optimizer and the stochastic gradient descent (SGD) optimizer, respectively, and the best parameters for the model were selected by mIoU. The comparison results are listed in Table 5. The quantitative analysis shows that the ELKPPNet prediction is better than all of the state-of-the-art methods in all four metrics.

Table 5. Quantitive analysis on 13-class NYUDv2 dataset (unit: %).

| Metric / Model | mIoU | FWIoU | PixelAcc | MeanClassAcc |
|---|---|---|---|---|
| Deeplabv3 | 28.23 | 38.62 | 54.47 | 46.12 |
| Deeplabv3+ | 40.15 | 51.21 | 66.83 | 52.47 |
| DenseASPP | 34.46 | 45.91 | 61.80 | 46.42 |
| PSPNet | 26.57 | 38.17 | 54.46 | 38.00 |
| RefineNet | 38.51 | 48.61 | 64.76 | 51.51 |
| Our ELKPPNet (parallel) | 45.28 | 54.42 | 69.81 | 58.43 |

**37-class NYUDv2 dataset:** The 37-class dataset is derived from the largest indoor scene parsing benchmark SUN-RGBD (Xiao et al. 2013), which includes B3DO (Janoch et al. 2013), NYUDv2, and other datasets. Because the number of classes is so large that it will occupy much GPU memory and result in the out of memory problem with a big batch size, we resized all the images to 384×288. In addition, in terms of the unbalanced occurrence frequencies cause by the highly varied categories, we added 500 more images from the test set into the training set, to learn more features. The scores of all the models are listed below.

Table 6. Quantitive analysis on 37-class NYUDv2 dataset (unit: %).

| Metric / Model | mIoU | FWIoU | PixelAcc | MeanClassAcc |
|---|---|---|---|---|
| Deeplabv3 | 28.51 | 48.32 | 64.27 | 34.48 |
| Deeplabv3+ | 29.30 | 50.09 | 65.69 | 35.03 |
| DenseASPP | 30.77 | 50.53 | 67.13 | 35.36 |
| PSPNet | 24.11 | 45.75 | 61.18 | 29.93 |
| RefineNet | 29.40 | 50.79 | 66.92 | 34.43 |
| Our ELKPPNet (parallel) | 34.41 | 55.11 | 70.03 | 39.00 |

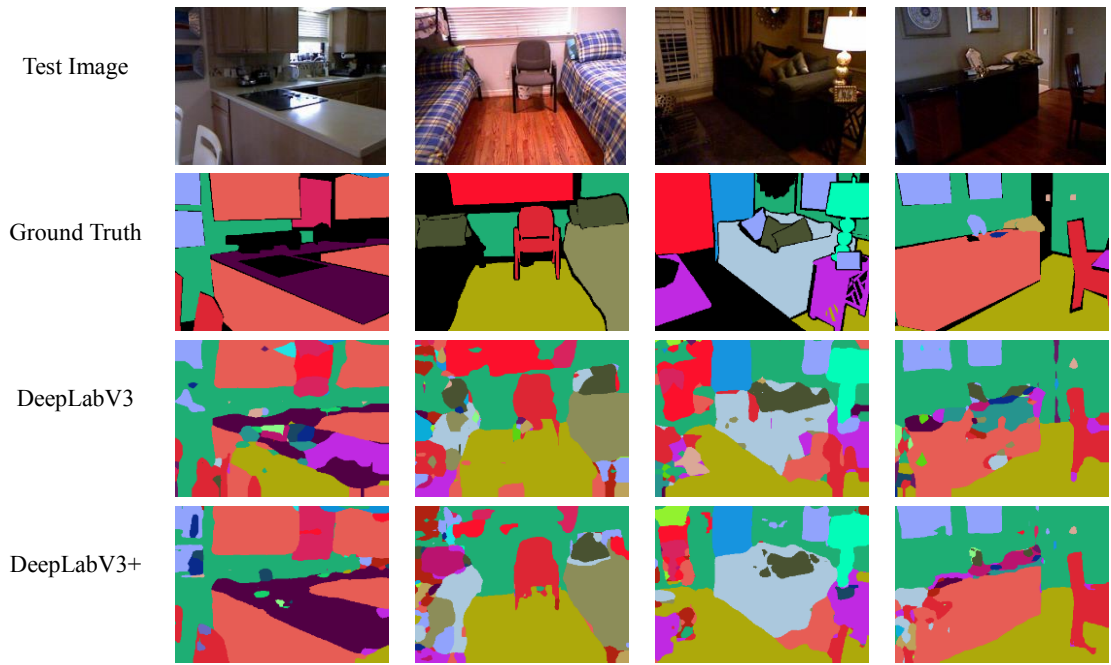

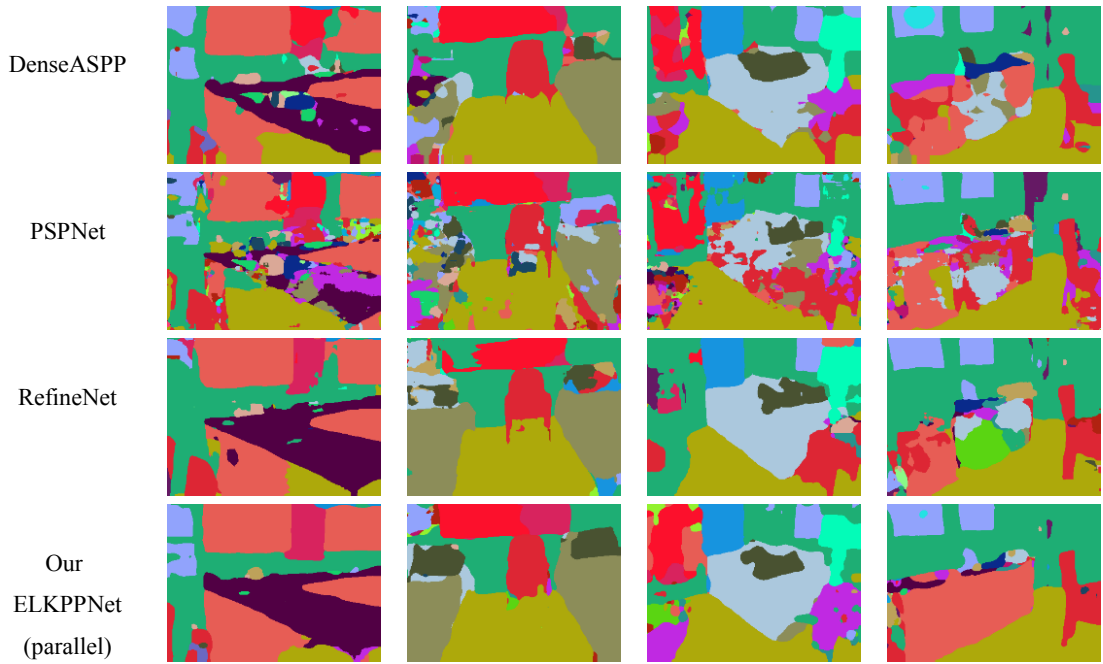

Fig. 11.Qualitative results on 37-class NYUDv2 dataset.

On the NYUDv2 dataset, although ELKPPNet outperforms the other methods in terms of all the metrics, all the deep learning models show a poor performance. This can be attributed to four reasons. Firstly, few training tricks are used, especially tricks such as random scaling and cropping, which may significantly improve multi-scale object detection and detail mining. Secondly, the void class of NYUDv2 covers a larger area than that of most outdoor datasets, and it even occupies nearly 50% in some images. Thirdly, there are frequent occlusions and severe illumination variation, impeding object feature recognition. Finally, all the models, including ELKPPNet, are not specifically designed for indoor scene parsing, where models for such a situation usually contain a well-designed structure for merging depth information (Gupta et al. 2015; Lin et al. 2017; Park et al. 2017). Moreover, in the 13-class dataset, many objects are categorized as the same, but they are indeed characterized by quite different features and geometrical structures; for example, the people, the cup, and the computer are all labeled as the object class.

## 5 Conclusion

In this paper, we have presented ELKPPNet for learning discriminative features in semantic segmentation. The proposed ELKPPNet model employs two simple yet effective convolutional operations for improving the semantic segmentation and a new edge-aware loss function for detail refinement. The large kernel pyramid pooling (LKPP) module is designed to encode multi-scale information with gradually increasing field of views and capture long-range context, while the balanced encoder-decoder framework is built for multi-level feature aggregation. The edge-aware cross-entropy (ECE) loss function focuses on tackling intra-class inconsistency and inter-class indistinction, and is able to capture unlabeled objects with an edge map, which means robustness to low-quality data (having unannotated objects). The ablation studies conducted confirmed the advantage of the LKPP module, the balanced encoder-decoder structure, and the ECE loss function in discriminative feature learning. The experimental results also showed that ELKPPNet exceeds the performance of the state-of-the-art models, under conditions with no special training tricks.

In our future work, we will focus on detail mining to improve the semantic segmentation quality,

especially for indoor scenes. We also hope to design a loss function for detail refinement with less dependence on hyper-parameters. Since the proposed edge-aware loss function can detect unlabeled objects with only edge information, it will be possible to explore this loss function's ability to perform few-shot learning in the future.